\documentclass[runningheads]{llncs}

\usepackage{accv}

\usepackage[dvipsnames]{xcolor}

\definecolor{gray}{rgb}{0.3,0.3,0.3}
\definecolor{blue}{rgb}{0,0.5,1}
\definecolor{mask_red}{rgb}{1,0,0.8}
\definecolor{green}{rgb}{0.2,1,0.2}
\definecolor{rblue}{rgb}{0,0,1}
\definecolor{lightblue}{HTML}{6495ed}
\definecolor{lightred}{HTML}{F19C99}

\newcommand{\green}[1]{\textcolor[RGB]{96,177,87}{#1}}

\newcommand{\fn}[1]{\footnotesize{#1}}
\newcommand{\gbf}[1]{\green{\bf{\fn{(#1)}}}}

\newcommand{\obf}[1]{\textcolor{orange}{\bf{\fn{(#1)}}}}
\definecolor{graytablerow}{gray}{0.6}

\definecolor{rblue}{rgb}{0,0.5,1}

\usepackage[width=122mm,left=12mm,paperwidth=146mm,height=193mm,top=12mm,paperheight=217mm]{geometry}

\usepackage{tikz}

\makeatletter
\DeclareRobustCommand\onedot{\futurelet\@let@token\@onedot}
\def\@onedot{\ifx\@let@token.\else.\null\fi\xspace}

\def\ie{\emph{i.e}\onedot}

\makeatother

\usepackage{graphicx}
\usepackage{booktabs}
\usepackage{amsmath}
\usepackage{amssymb}
\usepackage{xcolor}
\usepackage{color}
\usepackage{bbding}
\usepackage{array,multirow,textcomp}
\usepackage{rotating}
\usepackage{icomma} 
\usepackage[skip=0.5\baselineskip]{caption}
\usepackage{makecell}
\usepackage{booktabs}
\usepackage{adjustbox}
\usepackage{array}
\usepackage{balance}
\usepackage{wrapfig}
\usepackage{enumitem}
\usepackage{paralist}
\usepackage{tikz}
\usepackage{tabularx}
\usepackage{colortbl}
\usepackage{placeins}
\usepackage{float}
\usepackage{pgf}
\usepackage{subcaption}

\usepackage[accsupp]{axessibility}  
\usepackage[pagebackref,breaklinks,colorlinks,citecolor=blue,linkcolor=red]{hyperref}

\begin{document}
\pagestyle{headings}
\mainmatter

\def\ACCV24SubNumber{***}

\title{OneBEV: Using One Panoramic Image for Bird’s-Eye-View Semantic Mapping}
\titlerunning{OneBEV}
\authorrunning{J. Wei \emph{et al.}}

\author{Jiale Wei \and Junwei Zheng \and Ruiping Liu \and Jie Hu \and \\ Jiaming Zhang\thanks{Corresponding: jiaming.zhang@kit.edu} \and Rainer Stiefelhagen}

\institute{Karlsruhe Institute of Technology, Germany\\
\url{https://github.com/JialeWei/OneBEV}
}

\maketitle

\begin{abstract}
\sloppy
In the field of autonomous driving, Bird's-Eye-View (BEV) perception has attracted increasing attention in the community since it provides more comprehensive information compared with pinhole front-view images and panoramas. Traditional BEV methods, which rely on multiple narrow-field cameras and complex pose estimations, often face calibration and synchronization issues. To break the wall of the aforementioned challenges, in this work, we introduce OneBEV, a novel BEV semantic mapping approach using merely a single panoramic image as input, simplifying the mapping process and reducing computational complexities. A distortion-aware module termed Mamba View Transformation (MVT) is specifically designed to handle the spatial distortions in panoramas, transforming front-view features into BEV features without leveraging traditional attention mechanisms. Apart from the efficient framework, we contribute two datasets, \ie, nuScenes-360 and DeepAccident-360, tailored for the OneBEV task. Experimental results showcase that OneBEV achieves state-of-the-art performance with $51.1\%$ and $36.1\%$ mIoU on nuScenes-360 and DeepAccident-360, respectively. This work advances BEV semantic mapping in autonomous driving, paving the way for more advanced and reliable autonomous systems.
\end{abstract}
\section{Introduction}
\label{sec:intro}

\begin{figure}[t]
\centering
\includegraphics[width=\textwidth]{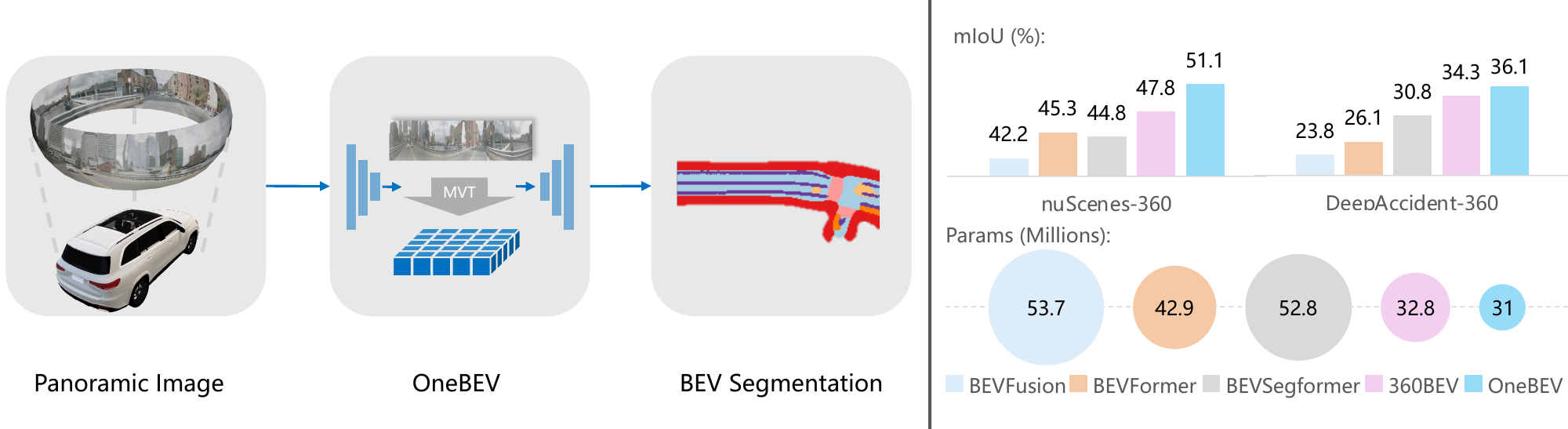}
\caption{
\textbf{OneBEV} only requires one panoramic image for BEV semantic mapping, decoupling the specifications of the camera, and can be more flexibly applied to different autonomous vehicle platforms. Achieving higher performance (mIoU) with fewer parameters on both nuScenes-360 and DeepAccident-360 shows the efficacy of OneBEV. 
}
\label{fig:fig1}
\vspace{-20pt}
\end{figure}

Autonomous driving solutions that rely solely on cameras are becoming mainstream and showing promising results~\cite{roddick2020predicting}. The bird's-eye view is a natural and straightforward candidate to serve as a unified representation~\cite{li2023delving}. Since bird's-eye view provides holistic information of locations and scale of objects, it is widely used in various autonomous driving scenarios such as perception and planning~\cite{bevformer,li2023delving}. Current BEV techniques~\cite{peng2023bevsegformer,liu2023bevfusion,ge2023metabev,wang2023unitr,borse2023xalign} rely on multiple narrow-field cameras to capture different views, then transform these front-view perspectives into a top-view and combine them for a comprehensive BEV representation. Nevertheless, these methods are constrained by their reliance on intrinsic and extrinsic camera parameters for pose estimation and view mapping. Calibration imperfections, synchronization difficulties, environmental conditions can cause discontinuity and inconsistency in scene perception, resulting in poor BEV features and unsatisfactory performance. Besides, the previous methods~\cite{peng2023bevsegformer,liu2023bevfusion,ge2023metabev,wang2023unitr,borse2023xalign} often suffer from higher computational complexity due to the requirement of processing data from multiple viewpoints in a scene.

To address the above issues, we propose \textbf{OneBEV}, a novel task to utilize \textbf{One} panorama for \textbf{BEV} semantic mapping in outdoor scenes. As shown in Fig.~\ref{fig:fig1}, this paradigm requires a single 360{\textdegree} image as input and gives a dense semantic map as output. The benefits of OneBEV include: (1) consistent omni-directional perception, (2) simplified camera setup, (3) no need for camera synchronization, and (4) lower computational complexity. 

However, the use of panoramas is always accompanied by spatial distortion and object deformation~\cite{zhang2022bending}. In order to solve this problem, we introduce a novel module Mamba View Transformation (MVT), a key component of OneBEV model. Inspired by the recent success of Mamba~\cite{gu2023mamba,liu2024vmamba,zhu2024vision} in text sequence and image modeling, we incorporate the scanning mechanism from Mamba to tackle the transformation from front-view panoramic perspective to BEV. Additionally, MVT integrates deformability capabilities, enhancing its suitability for overcoming distortions compared to using pure Mamba. Building on these innovations, the OneBEV architecture consists of three main components: the feature encoder, the view transformation module, and the semantic decoder. We use the VMamba-T~\cite{liu2024vmamba} model as the backbone for feature extraction. The MVT module then selects and transforms these features into BEV space. Finally, a lightweight semantic decoder processes these BEV features to generate the final semantic segmentation map.

Furthermore, the datasets used are crucial for the success of OneBEV. But no dataset is available in the literature for OneBEV task. In order to overcome the lack of training data, we introduce two OneBEV datasets, nuScenes-360 and DeepAccident-360. These datasets are expansions of the well-known nuScenes~\cite{caesar2020nuscenes} and DeepAccident~\cite{wang2024deepaccident} datasets, which are highly regarded in the field of autonomous driving. The nuScenes dataset provides diverse urban driving data, including images from six views and the semantic maps. We carefully combine images from the six perspectives into a unified panoramic image, aligning them using intrinsic and extrinsic camera parameters (these parameters are only used for dataset construction, not for training).  Similarly, the DeepAccident dataset, generated with the CARLA~\cite{dosovitskiy2017carla} simulator, offers synthetic yet realistic accident scenarios. We apply the consistent process to create panoramic images from its multi-view setup. In the end, for nuScenes-360 and DeepAccident-360, we contribute $34$K and $48$K panoramas with annotations, respectively. 

Extensive experiments prove that OneBEV achieves state-of-the-art performance with $51.1\%$ and $36.1\%$ mIoU on nuScenes-360 and DeepAccident-360, respectively, and has $1.8$ million fewer parameters than the baseline. OneBEV provides a practical solution for semantic scene understanding in dynamic outdoor environments by simplifying the process and minimizing computational complexity. This progress is noteworthy for the development of more efficient and dependable autonomous driving systems.

To summarize, our contributions can be outlined as follows: 
\begin{compactitem}
\item We introduce a new \textbf{OneBEV} paradigm, performing Bird's-Eye-View semantic mapping by using one 360\textdegree~image. 
\item To address data scarcity, we introduce \textbf{nuScenes-360} and \textbf{DeepAccident-360} datasets for the first time. We benchmark OneBEV by comparing multi-view methods on both real and synthetic datasets.
\item A new \textbf{Mamba View Transformation (MVT)} module is proposed to achieve efficient BEV transformation by using State Space Model (SSM) instead of cross-attention mechanism.
\item Extensive experiments and ablation studies prove the effectiveness of OneBEV paradigm and architecture, paving the way for new BEV semantic mapping.
\end{compactitem}

\section{Related Work}
\label{sec:related_work}
\subsection{State Space Models}
\label{sec:ssm}
State Space Models (SSMs)~\cite{gu2021efficiently}, like Transformers and RNNs, are used to process sequential data such as text, signals, etc. Initially introduced in the S4~\cite{gu2021efficiently} model, SSMs offer a unique architecture capable of efficiently modeling global information. Based on S4, Mamba~\cite{gu2023mamba} enhances SSMs by adding a selection mechanism for time-varying parameters and hardware optimization to achieve efficient training and inference. Subsequently, inspired by the success of the Mamba model, VMamba~\cite{liu2024vmamba} and Vim~\cite{zhu2024vision} extend Mamba to visual tasks by proposing bidirectional scanning and cross-scanning mechanisms to achieve location-aware visual understanding. Pan-Mamba~\cite{he2024pan} and Sigma~\cite{wan2024sigma} transfer the successful experience of Mamba to solve more challenging multimodal learning tasks. In our work, we explore the ability of Mamba to extract features as a visual backbone and how to apply Mamba to the task of view transformation, which poses new challenges to the Mamba model in solving complex visual tasks.

\subsection{Panoramic Semantic Segmentation}
\label{sec:pano}
Semantic segmentation is a fundamental problem in the field of computer vision, and numerous mature methods have been developed~\cite{long2015fully,ronneberger2015u,zhao2017pyramid,guo2022segnext}. In recent years, panoramic semantic segmentation has garnered increasing attention~\cite{jaus2021panoramic,orhan2022semantic,xu2019semantic,yang2021context,zhang2022bending}. Unlike narrow-view pinhole camera images, panoramic images are typically obtained using equirectangular projection, which causes distortion and warping of objects in the image, posing challenges for semantic segmentation. In our work, we need to select appropriate datasets for our novel task. Datasets~\cite{ma2021densepass,liao2022kitti} focused on panoramic image perception tasks only contain semantic labels for front-view images and lack semantic labels from a BEV perspective. Moreover, panoramic images often lack camera parameters, making it impossible to convert front-view semantic labels to BEV perspective using homography transformation. Datasets that include multi-view pinhole camera images provide BEV perspective semantic labels, but the multi-view front images require additional processing to be stitched into panoramic images. This is the case with the nuScenes~\cite{caesar2020nuscenes} and DeepAccident~\cite{wang2024deepaccident} datasets used in our work. After our data processing, we contribute $34$K and $48$K panoramas with annotations, respectively. 

\subsection{BEV Semantic Mapping}
\label{sec:bev_sem}
For BEV semantic segmentation tasks, view transformation plays a core role~\cite{li2023delving}. It is mainly divided into two methods~\cite{li2023delving}: \textit{2D-3D} and \textit{3D-2D}. The \textit{2D-3D} method was first introduced by LSS~\cite{philion2020lift}, which estimates implicit pixel depth information from 2D features and then uses camera geometry to establish the connection between BEV segmentation and feature maps in 3D space. Additionally, BEVDet~\cite{huang2021bevdet}, M$^{2}$BEV~\cite{xie2022m}, and BEVFusion~\cite{liu2023bevfusion}, building on the work of LSS, have explored better depth estimation methods. The \textit{3D-2D} method was proposed many years ago by IPM~\cite{mallot1991inverse}, which uses homography transformation to convert camera images into BEV. Currently, mainstream works ~\cite{zhou2022cross,saha2022translating,bevformer,ge2023metabev,wang2023unitr} employ multi-layer perceptron (MLP) or Transformer architectures to perform 3D-to-2D projection using camera parameters. BEVSegformer~\cite{peng2023bevsegformer} completes the view projection without using camera parameters. In terms of panoramic views, 360BEV~\cite{teng2024360bev} was the first to apply the Transformer architecture to achieve view transformation using supervised ground-truth depth information. Based on 360BEV, we innovatively explore outdoor scenarios, and depth information doesn't included.

\section{OneBEV}
\label{sec:onebev}
\subsection{Task Definition}
\label{sec:task}
OneBEV is to crerate a BEV semantic map using only one panoramic image. This task streamlines the process of top-down perspective analysis across different domains, providing thorough spatial understanding using just one 360{\textdegree} image. To adapt to our new panoramic task, we combine together the multi-view images in the existing datasets~\cite{caesar2020nuscenes, wang2024deepaccident} to obtain the omni-directional panorama, which will be used as the \textbf{input} of the neural network. The high-definition semantic map in the dataset will be used as the \textbf{label} for network training. The entire process is optimized using focal loss as the \textbf{loss function}.

\begin{figure}[t]
\centering
\includegraphics[width=\textwidth]{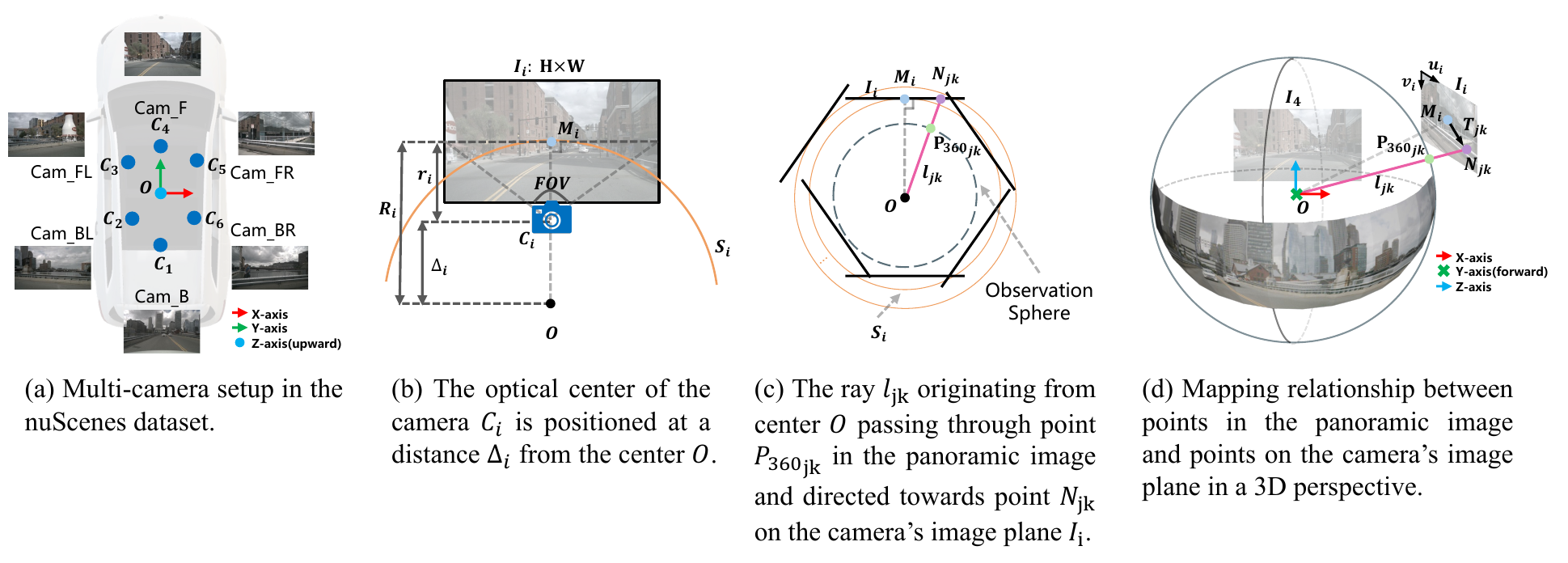}
\caption{\textbf{Data processing} of nuScenes-360 and DeepAccident-360 dataset. (a) The original multi-view setup requires complex camera configuration and higher computational complexity. (b) Given the front view as an example, the camera parameters are visualized in the original image plane. (c) The panoramic image is calculated via the observation sphere according to the camera image plane. (d) The definition of mapping relationship between narrow-FoV and omni-FoV images. 
}
\label{fig:pano_stitch}
\vspace{-20pt}
\end{figure}

\subsection{Datasets}
\label{sec:datasets}
\noindent\textbf{Data Processing.}
\label{sec:data_processing}
Based on the panoramic generation from 360BEV~\cite{teng2024360bev}, we project multi-view images onto a sphere following the order of 
\begin{center}
$C_1$:\textit{Camera\_Back}, $C_2$:\textit{Camera\_Back\_Left}, $C_3$:\textit{Camera\_Front\_Left}, $C_4$:\textit{Camera\_Front}, $C_5$:\textit{Camera\_Front\_Right}, $C_6$:\textit{Camera\_Back\_Right},
\end{center}

\noindent to obtain the panoramic image. The Fig.~\ref{fig:pano_stitch}(a) depicts the multi-camera setup on a vehicle in the nuScenes~\cite{caesar2020nuscenes} dataset, where six cameras are mounted on the car roof, with each camera rotated at an interval of 55{\textdegree}. All cameras possess a 70{\textdegree} field of view (FOV) except for \textit{Camera\_Back}, which has a 110{\textdegree} FOV. This configuration ensures a full 360{\textdegree} coverage with overlapping fields of view, facilitating effective stitching. Unlike indoor dataset setups in~\cite{teng2024360bev}, where cameras are mounted on a tripod and rotate around a single center in an indoor environment, the cameras in the autonomous driving dataset are mounted at different positions around the car.  Thus, as illustrated in Fig.~\ref{fig:pano_stitch}(a)(b), the optical centers $C_i$ $(i \in [1,6])$ of all cameras are distributed, indicating that each camera's imaging plane $I_i$ is tangential to spheres $S_i$ with different radii $R_i$, all centered around $O$, which is the average optical center of all cameras. 

In order to accurately stitch panoramic images, we begin by calculating the fundamental parameters for each camera (see Fig.~\ref{fig:pano_stitch}(b)). Initially, we determine the distance $r_i$ from the optical center $C_i$ of each camera to its imaging plane $I_i$ by utilizing the image width $W$ and FOV. Next, we calculate $\Delta_i$, which represents the displacement of $C_i$ from $O$. It is crucial to mention that $\Delta_{i_x}$ and $\Delta_{i_y}$ represent displacements along the $x$-axis and $y$-axis, respectively, and are measured in millimeters. In order to align $r_i$, which is given in pixel coordinates, we need to convert $\Delta_i$ to the pixel coordinate system. The pixel size of each pixel on the $x$-axis and $y$-axis is determined by the real camera focal length $f$ and the values $f_x$ and $f_y$ in the camera intrinsic matrix $\mathbf{K}$. This relationship is expressed by the following equation:
\begin{equation}
f_x, f_y[pixel] = \frac{f[mm]}{pixel\_size}.
\end{equation}
We proceed by converting $\Delta_{i_x}$ and $\Delta_{i_y}$ to pixel units respectively in order to obtain $\Delta_i$. This conversion is done using a series of equations:
\begin{equation}
\begin{aligned}
\Delta_{i_u}[pixel] & = \Delta_{i_x}[mm]*pixel\_size = \Delta_{i_x}[mm]\frac{f_x[pixel]}{f[mm]},   \\
\Delta_{i_v}[pixel] & = \Delta_{i_y}[mm]*pixel\_size = \Delta_{i_y}[mm]\frac{f_y[pixel]}{f[mm]},   \\
\Delta_i[pixel] & = \sqrt{{\Delta_{i_u}}^2+ {\Delta_{i_v}}^2} = \sqrt{{(\Delta_{i_x}\frac{f_x}{f})}^2+ {(\Delta_{i_y}\frac{f_y}{f})}^2}\\
& = \sqrt{{((C_{i_{x}}-O_x)\frac{f_x}{f})}^2+ {((C_{i_{y}}-O_y)\frac{f_y}{f})}^2}.
\end{aligned}
\end{equation}

\noindent Finally, the radius $R_i$ of the sphere $S_i$ corresponding to each camera's imaging plane $I_i$ can be obtained via the following formula:
\begin{equation}
R_i = r_i + \Delta_i.
\end{equation}

\begin{figure}[]
\vspace{-20pt}
\centering
\includegraphics[width=0.8\textwidth]{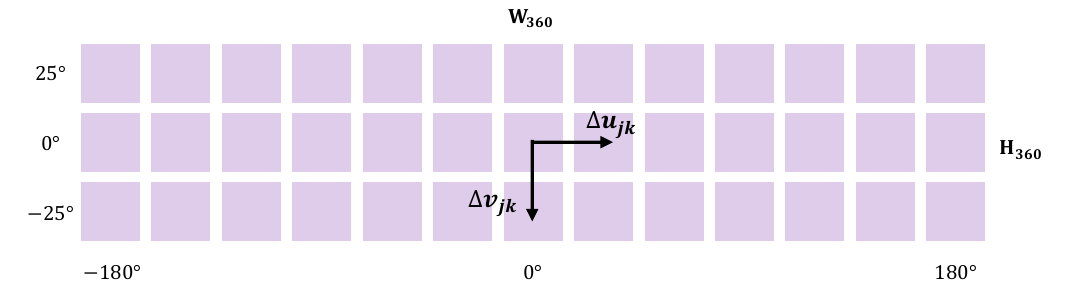}
\caption{\textbf{Pixel mapping relationship in panoramas}. Each pixel in the panoramic image is mapped to angular deviations from the central pixel in the horizontal and vertical directions.}
\label{fig:angular_deviations}
\vspace{-20pt}
\end{figure}

As illustrated in Fig.~\ref{fig:pano_stitch}(d), the final panoramic image is projected onto a observation sphere, which, when unfolded, yields a 2D panoramic image $f_{360} \in \mathbb{R}^{H_{360}\times W_{360}\times 3}$. Each pixel $P_{{360}_{jk}}$ $(j \in [1,H_{360}], k \in [1,W_{360}])$ in the 2D panoramic image is mapped horizontally $\pm$180{\textdegree} and vertically $\pm$25{\textdegree} based on the FOV. This mapping allows us to determine the angular deviations $\Delta u_{jk}$ and $\Delta v_{jk}$ between each pixel and the center of the 2D image (see Fig.~\ref{fig:angular_deviations}). By utilizing the radius $R_i$ and angular deviations, we determine two crucial points $M_{i}$ and $N_{jk}$ during the stitching procedure (see Fig.~\ref{fig:pano_stitch}(c)(d)). These points are used to establish the mapping between points in the panoramic image and points on the camera’s image plane.

$M_{i}$:
Each camera’s image plane $I_{i}$ is tangent to its corresponding sphere $S_{i}$ at point $M_{i}$. Therefore, the image plane $I_{i}$ can be mathematically represented by Eq.~\ref{eq:I}. First, we compute the orientation offset $[\Delta yaw_i, \Delta pitch_i, \Delta roll_i]$ between the camera $C_i$ and the reference camera $C_4$ using the camera extrinsic parameters. Then, we transform this offset from Euler angles to the Cartesian coordinates of point $M_{i}$ on the surface of the obeservation sphere using Eq.~\ref{eq:m}:
\begin{equation}\label{eq:I}
\begin{array}{c}
M_{i_x} (x - M_{i_x}) + M_{i_y}(y - M_{i_y}) + M_{i_z}(z - M_{i_z}) = 0,\\
M_{i_x}^{2} + M_{i_y}^{2} + M_{i_z}^{2} = R_i^2,
\end{array}
\end{equation}

\begin{equation}\label{eq:m}
M_i = \begin{bmatrix} M_{i_x}\\M_{i_y} \\ M_{i_z}\end{bmatrix}
=\begin{bmatrix}  
  R_i * \cos(\Delta pitch_i) * \sin(\Delta yaw_i) \\  
  R_i * \cos(\Delta pitch_i) * \cos(\Delta yaw_i)  \\  
  R_i * \sin(\Delta pitch_i) 
\end{bmatrix}.  
\end{equation}

$N_{jk}$:
The view line $l_{jk}$ is represented by a ray that originates from the center of the observation sphere $O$, passes through a point $P_{{360}_{jk}}$ in the panoramic image, and intersects the camera's image plane $I_{i}$ at point $N_{jk}$. Eq.~\ref{eq:l} describes the formula of the view line:
\begin{equation}\label{eq:l}
\begin{aligned}
\begin{matrix} 
x = \alpha_{jk} t \\
y = \beta_{jk} t \\
z = \gamma_{jk} t \\
\end{matrix}\implies \frac{x}{\alpha_{jk}} = \frac{y}{\beta_{jk}}= \frac{z}{\gamma_{jk}},
\end{aligned}
\end{equation}
where $x$, $y$, and $z$ are proportional to $t$ with proportionality constants $\alpha_{jk}$, $\beta_{jk}$, and $\gamma_{jk}$. We solve the equations of the image plane (in Eq.~\ref{eq:I}) and the view line (in Eq.~\ref{eq:l}) by utilizing the angular deviations $\Delta u_{jk}$ and $\Delta v_{jk}$ in order to determine $N_{jk}$ (in Eq.~\ref{eq:n}):
\begin{equation}\label{eq:n}
\begin{aligned}
\begin{bmatrix}
\alpha_{jk}\\\beta_{jk}\\\gamma_{jk}\\
\end{bmatrix} & = \begin{bmatrix}\cos(\Delta v_{jk})\sin(\Delta u_{jk})\\
\cos(\Delta v_{jk})\cos(\Delta u_{jk})\\
 \sin(\Delta v_{jk}), \\
\end{bmatrix},\\
t & = \frac{R_i^2}{M_{i_x}\alpha_{jk} + M_{i_y}\beta_{jk} + M_{i_y}\gamma_{jk}} ,\\
N_{jk} & = \begin{bmatrix} N_{{jk}_x}\\N_{{jk}_y} \\ N_{{jk}_z}\end{bmatrix} = \begin{bmatrix}  
  \alpha_{jk} t \\  
  \beta_{jk} t  \\  
  \gamma_{jk} t 
\end{bmatrix}.
\end{aligned}
\end{equation}

The coordinates of $N_{jk}$, initially calculated in the observation sphere’s coordinate system, must be transformed into the image plane’s coordinate system. We know that $M_{i}$ lies on the image plane. The vector from $M_{i}$ to $N_{jk}$ is denoted as $T_{jk}$. We first define the unit vector in the positive $u$ and $v$ directions:
\begin{equation}\label{eq:uv}
\begin{aligned}
u_i & = [\cos(\Delta yaw_i), -\sin(\Delta yaw_i), 0],\\
v_i & = M_i \times u_i.
\end{aligned}
\end{equation}
By projecting the vector $T_{jk}$ onto the vectors $u_i$ and $v_i$, we can determine the displacement of $N_{jk}$ with respect to $M_{i}$ on the image plane $I_i$. This displacement is represented by the values $\Delta u_{jk}$ and $\Delta v_{jk}$.

\begin{figure}[]
\centering
\vspace{-4pt}
\includegraphics[width=0.95\textwidth]{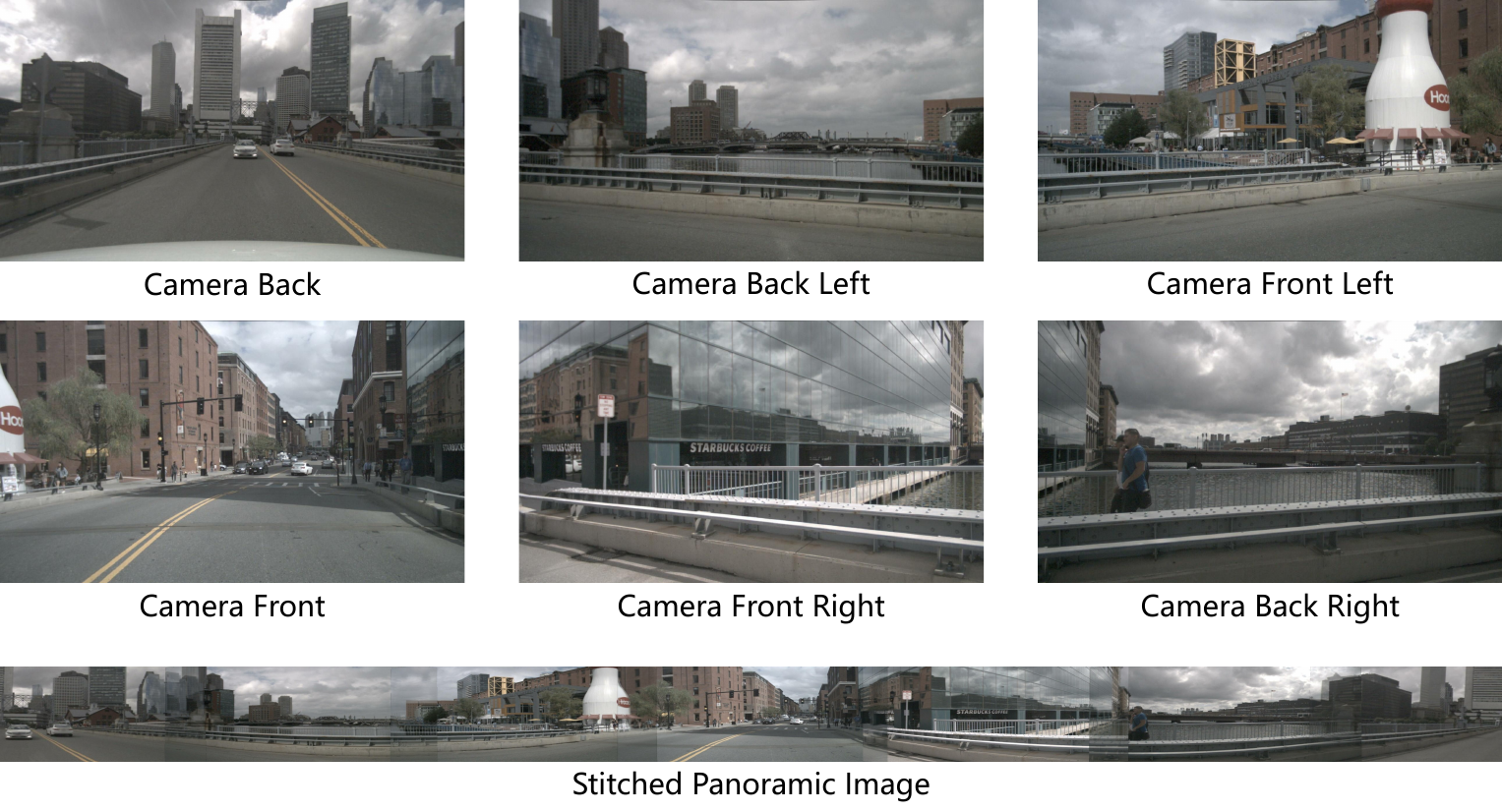}
\caption{\textbf{Visualization of data samples}. The first two rows are the multi-view images from the original nuScenes dataset, while the last row is the reconstructed panorama. }
\label{fig:fv_pano}
\vspace{-20pt}
\end{figure}

These displacements allow us to locate $N_{jk}$ on the image plane, thereby determining the corresponding pixel in the original camera image. If the displacement values fall outside the image boundaries, the point is considered invalid. By repeating this process for every point in the panoramic image, we construct the final panorama with contributions from all cameras, ensuring complete coverage. The example of multi-view images and the panoramic image are shown in Fig.~\ref{fig:fv_pano}.

\noindent\textbf{NuScenes-360 and DeepAccident-360 Dataset.}
The original nuScenes~\cite{caesar2020nuscenes} dataset is a large-scale outdoor dataset in the field of autonomous driving. It includes 6 narrow-view images, LiDAR sweeps, Radar sweeps, semantic maps, and other features collected during driving in various urban environments. We combine the multi-view images using the camera parameters provided in the dataset and extracted the required classes from the high-definition semantic maps as labels, forming the new nuScenes-360 dataset. To explore different scenarios, we also introduce another synthetic dataset DeepAccident~\cite{wang2024deepaccident}, which is generated from CARLA~\cite{dosovitskiy2017carla} Simulator. DeepAccident has a similar structure to nuScenes, thus we employ the same generation to create new DeepAccident-360 dataset.

\begin{figure}[]
\centering
\includegraphics[width=\linewidth]{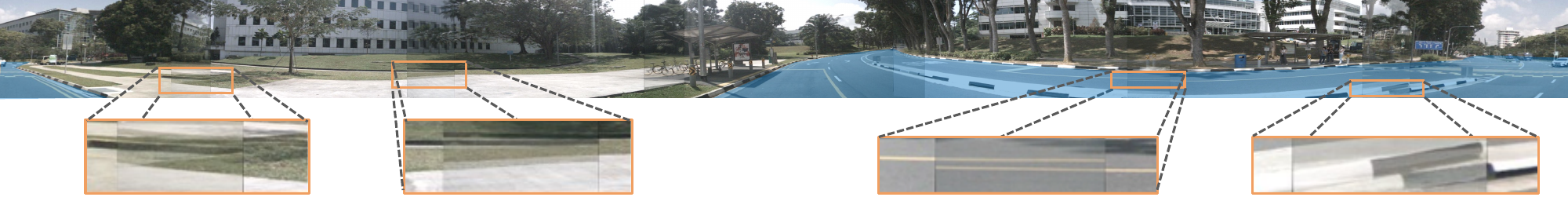}
\caption{\textbf{Challenges in panoramic images}. For example, a straight road appears curved and incomplete in the image (\textcolor{blue}{blue shade}), distributed across different positions. Due to stitching quality issues, seams and ghosting can appear between the original adjacent images (\textcolor{orange}{orange boxes}), causing misalignment between objects.}
\label{fig:challenges}
\vspace{-3pt}
\end{figure}

\noindent\textbf{OneBEV Challenges.}
Since the overlapping regions between multiple cameras in datasets are limited, there are cases where the same object does not appear in adjacent images. Therefore, using feature points based methods~\cite{lowe2004distinctive,bay2006surf,rublee2011orb} for matching can be slow or even fail. Our stitching method based on the physical model of the camera, which, although fast, does not achieve high stitching quality. Additionally, objects in the panoramic image may become distorted or warped, presenting challenges to the robustness of the neural network. Fig.~\ref{fig:challenges} illustrates some of these challenges.

\begin{figure}[t]
    \centering
    \begin{minipage}[t]{0.46\linewidth}
        \centering
        \frame{\includegraphics[width=\linewidth]{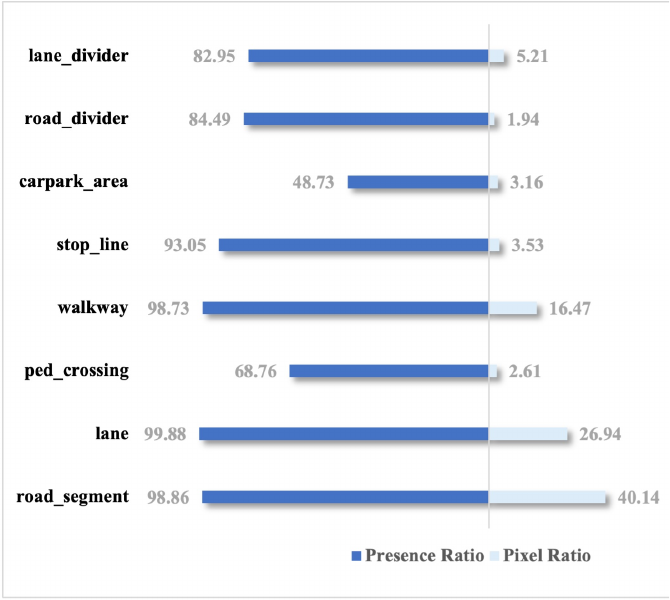}}
        \subcaption{Distribution  of nuScenes dataset.}
        \label{fig:dataset_stat_a}
    \end{minipage}%
    \hfill
    \begin{minipage}[t]{0.46\linewidth}
        \centering
        \frame{\includegraphics[width=\linewidth]{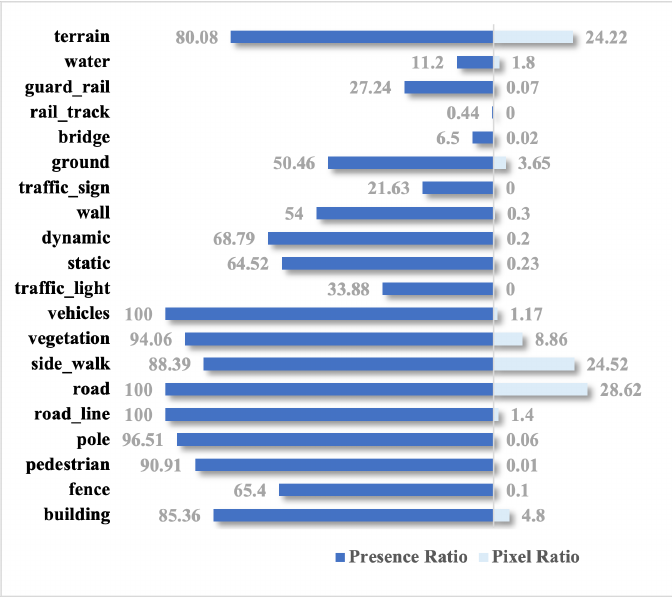}}
        \subcaption{Distribution of DeepAccident dataset.}
        \label{fig:dataset_stat_b}
    \end{minipage}
    \caption{\textbf{Analysis of semantic class distributions}. The per-class pixel ratio and presence ratio are presented in two datasets. Pixel ratio (\%): Pixels of each class across all frames. Presence ratio (\%): Frames where a class appears (at least 2 pixels).}
    \label{fig:dataset_stat}
    \vspace{-20pt}
\end{figure}

\noindent\textbf{OneBEV Dataset Statistics.}
We accomplish an analysis of the semantic classes present in the dataset and carefully choose the appropriate ones to be used for training purposes. Two measures, namely ``\textit{pixel ratio}'' and ``\textit{presence ratio}'', are employed for evaluating these semantic classes. Specifically, the pixel ratio is the total number of pixels for each class across all frames divided by the total number of pixels for all classes across all frames. Similarly, the presence ratio is defined as the ratio of the total number of frames in which a particular class is present (considered present if at least two pixels belong to that class) to the total number of frames. The pixel ratio and presence ratio of each class in the nuScenes~\cite{caesar2020nuscenes} and DeepAccident~\cite{wang2024deepaccident} datasets are shown in Fig.~\ref{fig:dataset_stat}. For the nuScenes-360 dataset, we retain or merge the classes (Fig.~\ref{fig:dataset_stat_a}) from the original nuScenes dataset based on the settings in~\cite{liu2023bevfusion}, resulting in six classes:  \textit{drivable\_area(road\_segment, lane)}, \textit{ped\_crossing}, \textit{walkway}, \textit{stop\_line}, \textit{carpark\_area}, and \textit{divider(road\_divider, lane\_divider)}. For the DeepAccident-360 dataset, we analyze the 23 original classes. Initially, we remove the labels \textit{unlabeled}, \textit{other}, and \textit{sky} as they are not relevant to our work. Based on the statistics in Fig.~\ref{fig:dataset_stat_b}, we further eliminate labels with a pixel ratio of 0: \textit{traffic\_sign}, \textit{traffic\_light}, and \textit{rail\_track}, whose presence ratios are also less than 50\%. This process ultimately results in 17 valid classes. The final panoramic datasets statistics are shown in Table~\ref{tab:datasets}.

\begin{table}[h]
\vskip -2ex
\begin{center}
\caption{\textbf{Data statistics} of nuScenes-360 dataset and DeepAccident-360 dataset.}
\vskip -2ex
\label{tab:datasets}
\setlength{\tabcolsep}{6mm}
\renewcommand{\arraystretch}{0.9}
\resizebox{0.9\columnwidth}{!}{
\begin{tabular}{lcrr}
\toprule
\textbf{Dataset} & \textbf{\#Scene} & \textbf{\#Frame} & \textbf{\#Class} \\
\midrule
nuScenes-360 \texttt{train} & 700 & 28,130 & 6\\
nuScenes-360 \texttt{val}   & 150  & 6,019 & 6 \\
nuScenes-360 \texttt{total} & 850 & 34,149 & 6 \\ \midrule
DeepAccident-360 \texttt{train} & 483 & 40,619 & 17\\
DeepAccident-360 \texttt{val} & 104 & 8,193 & 17\\
DeepAccident-360 \texttt{total}& 587 & 48,812 & 17 \\
\bottomrule
\end{tabular}
}    
\end{center}
\vskip -3ex
\end{table}

\subsection{Model}

\begin{figure}[t]
\centering
\includegraphics[width=\linewidth]{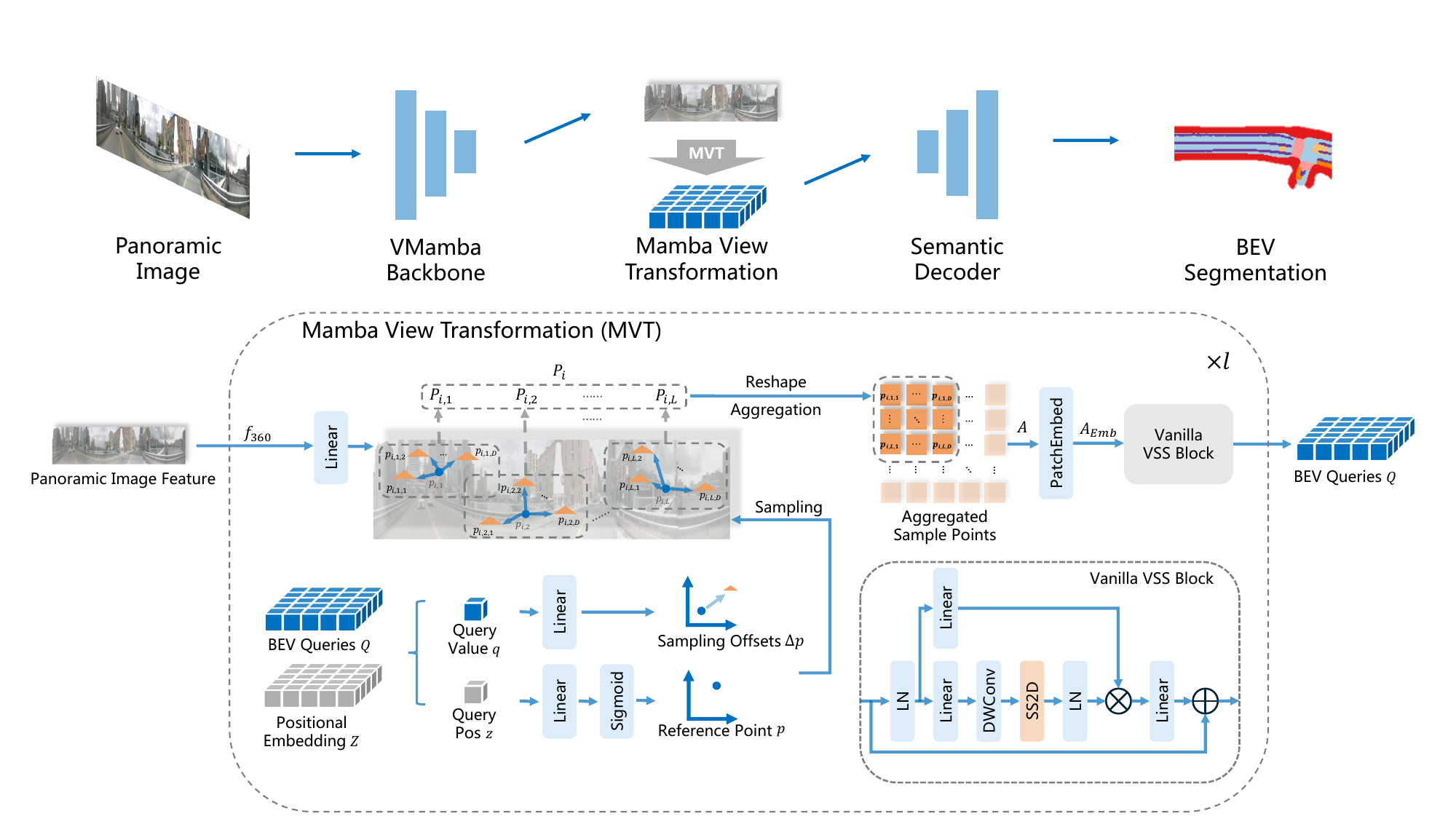}
\caption{\textbf{Overall architecture of the proposed OneBEV.} The VMamba-T~\cite{liu2024vmamba} model as the backbone extracts features from the panorama. The Mamba-based view transformation (MVT) module transforms the front view features into bird’s eye view features. A lightweight semantic decoder supports the semantic segmentation task.}
\label{fig:model_onebev}
\vspace{-20pt}
\end{figure}

\noindent\textbf{OneBEV Architecture.}
OneBEV is an end-to-end model focusing on BEV semantic segmentation using panoramic image. Fig.~\ref{fig:model_onebev} illustrates the overall structure. The front-view panoramic image serves as the input for the OneBEV model. The VMamba-T~\cite{liu2024vmamba} backbone is employed to extract features from this input, ensuring that high-quality and relevant information is captured. Following this, the MVT module plays a crucial role by selecting specific feature points on the front-view image to be used as BEV queries. These selected feature points are then sent into the vanilla VSS block~\cite{liu2024vmamba} for interaction and processing. In the final stage, the semantic segmentation head takes charge, utilizing the processed features to perform the task of semantic segmentation.

\noindent\textbf{Mamba View Transformation (MVT).}
The input to our OneBEV model is a panoramic image, and it lacks camera intrinsic and extrinsic parameters. The spatial cross-attention mechanism in BEVFormer~\cite{bevformer} accomplishes view transformation based on these parameters, which is not applicable to our task. We design a novel module akin to the multi-camera deformable cross-attention mechanism~\cite{peng2023bevsegformer} between the front-view feature map and the BEV feature map, termed Mamba View Transformation (MVT). The detailed structure is depicted in the lower half of Fig.~\ref{fig:model_onebev}.

Following common practices, we first define BEV queries $Q \in \mathbb{R}^{H_q \times W_q \times C_{Emb}}$, along with their positional embeddings $Z \in \mathbb{R}^{H_q \times W_q \times C_{Emb}}$. For BEV queries $Q$, we perform a linear projection to obtain the sampling offsets $\Delta p \in \mathbb{R}^{H_q \times W_q \times N \times 2}$, while its positional embedding $Z$ undergoes a linear projection followed by normalization through a sigmoid function to derive the reference points $p \in \mathbb{R}^{H_q \times W_q \times L \times 2}$, where $H_q$ and $W_q$ represent the spatial dimensions of BEV queries, $C_{Emb}$ is the number of embedded dimensions, $N$ is the total number of sampling points, and $L$ is the number of sampling locations (in our task, $N$ is always equal to $L^{2}$). The MVT module can be expressed as:
\begin{equation}\label{eq:2}
\begin{aligned}
P_{i} & = Wf_{360}(p_{i,j} + \Delta p_{i,j,k}), \\
A & = \text{Reshape}(P), \\
Q & = \text{VSS}(PatchEmbed(A)).
\end{aligned}
\end{equation}
Here, the panoramic features $f_{360} \in \mathbb{R}^{H_{360} \times W_{360} \times C_{Emb}}$ are linearly projected with the learnable parameter matrices $W \in \mathbb{R}^{C_{Emb} \times C_{Emb}}$. $p_{i,j}$ and $\Delta p_{i,j,k}$ denote the reference points and sampling offsets, respectively, where $i \in [1, ..., H_qW_q]$, $j \in [1, ..., L]$, $k \in [1, ..., D]$, and $D$ is the number of sampling points at each location (here $D=L$). Thus, we obtain a set of sampled points $P_{i}$ from the panoramic features for each query $q$. Subsequently, each set $P_{i} \in \mathbb{R}^{LD\times C_{Emb}}$ is reshaped into a grid form $P_{i} \in \mathbb{R}^{L\times D\times C_{Emb}}$, and the set of all $P_{i}$, denoted as $P \in \mathbb{R}^{LH_qDW_q\times C_{Emb}}$, is reshaped into a larger grid form $A \in \mathbb{R}^{LH_q\times DW_q\times C_{Emb}}$. This process can be understood as each BEV query point $q$ on the BEV feature map being ``interpolate'' by the sampling point grid $P_{i} \in \mathbb{R}^{L\times D\times C_{Emb}}$. The aggregated sample points $A$ can be interpreted as the BEV feature map $Q \in \mathbb{R}^{H_q \times W_q \times C_{Emb}}$ originally input into the MVT being ``interpolate'' into a larger feature map $A \in \mathbb{R}^{LH_q\times DW_q\times C_{Emb}}$. To ensure that the reshaped feature map $A \in \mathbb{R}^{LH_q\times DW_q\times C_{Emb}}$ can be effectively processed, we apply a patch embedding to convert it back to the original BEV feature map size $A_{Emb} \in \mathbb{R}^{H_q\times W_q\times C_{Emb}}$. Finally, $A_{Emb}$ is fed into vanilla VSS blocks from VMamba~\cite{liu2024vmamba} to enable interaction between feature points. $Q$ is the final output of the MVT module, which can then be used as the BEV queries of the next MVT module to perform sampling on the front view panoramic image.

Compared to the multi-camera deformable cross-attention module in BEVSegformer~\cite{peng2023bevsegformer}, our MVT module exhibits two main differences:
\textbf{(1)} Exclusive sampling on panoramic feature map: MVT operates solely on the panoramic feature map, rather than on multi-view images. This approach provides the network with enhanced flexibility in selecting reference points, allowing for more precise feature extraction.
\textbf{(2)} Absence of attention mechanism: Instead of the attention mechanism, our module introduces a novel Mamba-based method. In traditional attention mechanisms, attention weights are used to select sampled points relevant to each query. However, with the SS2D scanning operation in the vanilla VSS block, our approach not only facilitates interaction between each query's own sampled points but also enables interaction with sampled points from other queries, leading to a more integrated and comprehensive feature representation.

\section{Experiment}
\subsection{Experimental Settings}
\noindent\textbf{Implementation Details.} We set the embedding dimensions to $128$. For the MVT module, we configure it with $4$ layers, $5$ sampling locations, and set the inner dimensions and depth of the vanilla VSS block to $128$ and $2$, respectively. The OneBEV model is trained using $4$ A100 GPUs, with a batch size of $2$ per GPU. The initial learning rate is $5{\times}10^{-5}$ for the nuScenes-360 dataset and $8{\times}10^{-5}$ for the DeepAccident-360 dataset. The training schedule consists of $50$ epochs, utilizing both linear warm-up and cosine annealing strategy. We employ the AdamW optimizer with a weight decay of $0.01$. The input for OneBEV training comprises panorama images from both datasets, each sized at $600{\times}9600$. The output BEV map is set to $200{\times}200$, covering a perception range of $[-50m, 50m]\times [-50m, 50m]$ around the ego vehicle. Following~\cite{liu2023bevfusion}, we report the Intersection-over-Union (IoU) across $6$ classes for nuScenes-360 and $17$ classes for DeepAccident-360, along with the class-averaged mean IoU as our evaluation metrics.

\noindent\textbf{Baselines.}
We conduct a comparative analysis of OneBEV against previous state-of-the-art models (\textbf{BEVFormer}~\cite{bevformer}, \textbf{BEVSegformer}~\cite{peng2023bevsegformer}, \textbf{BEVFusion}~\cite{liu2023bevfusion}, \textbf{HDMapNet}~\cite{li2022hdmapnet}, \textbf{360BEV}~\cite{teng2024360bev}) on the BEV semantic segmentation task using nuScenes-360 and DeepAccident-360 datasets. All models except 360BEV~\cite{teng2024360bev} originally utilize images from multi-view cameras as input. In our work, we spent large effort to reproduce their models and adapt their narrow-FoV inputs to 360{\textdegree} panoramic images. For the 360BEV model, which inherently accepts panoramic images but requires ground-truth depth for transformation, we adapt it without using the depth for a fair comparison. To ensure fairness, we standardized their view transformation modules to the multi-camera deformable cross-attention from BEVSegformer, applied here with a single camera setup.

\subsection{Results}

\noindent\textbf{Results on Nuscenes-360 Dataset.}
Based on the results presented in Table~\ref{tab:result:nuscenes} and the comparative analysis provided, it is evident that our proposed method, OneBEV, demonstrates significant improvements in performance compared to 360BEV~\cite{teng2024360bev} on the nuScenes-360 dataset. Specifically, it shows a $+2.1\%$ improvement in \textit{Drivable Area} IoU, $+2.9\%$ in \textit{Pedestrian Crossing} IoU, $+3.7\%$ in \textit{Walkway} IoU, $+3.7\%$ in \textit{Stop Line} IoU, $+4.0\%$ in \textit{Carpark IoU}, $+2.4\%$ in \textit{Divider IoU} and $+3.3\%$ in all classes mIoU. These results indicate the effectiveness of OneBEV in semantic segmentation tasks, showcasing its robustness and adaptability across various classes in the nuScenes-360 dataset. The combination of superior accuracy and a reduced parameter count ($32.8$M $\rightarrow 31$M) positions OneBEV as a highly effective and efficient model for 360{\textdegree} BEV segmentation.
\begin{table*}[]
\vskip -4ex
\begin{center}
\setlength{\tabcolsep}{2pt}
\small\centering
\caption{Per-class results on the \texttt{val} set of nuScenes-360 dataset.}
\label{tab:result:nuscenes}
\vskip -2ex
\resizebox{0.99\textwidth}{!}{
\begin{tabular}{l|c|cccccc|l}
    \toprule
     Method & \#Param. & Drivable\_Area & Ped\_Crossing & Walkway & Stop\_Line & Carpark\_Area & Divider & Mean \\
    \midrule
    HDMapNet & 12.8M & 72.8 & 37.8 & 44.2 & 33.8 & 35.8 & 31.4 & 42.6 \\
    BEVFusion & 53.7M & 72.4 & 36.4 & 43.8 & 33.1 & 35.4 & 31.8 & 42.2 \\
    BEVSegformer & 52.8M & 73.6 & 40.3 & 45.4 & 36.2 & 40.5 & 32.6 & 44.8 \\
    BEVFormer & 42.9M & 74.7 & 40.5 & 46.1 & 38.1 & 37.3 & 35.0 & 45.3 \\
    360BEV & 32.8M& 77.1 & 44.7 & 49.5 & 39.0 & 40.3 & 36.6 & 47.8\\
    \textbf{OneBEV} (Ours) & \textbf{31M} & \textbf{79.2} & \textbf{47.6} & \textbf{53.2} & \textbf{43.4} & \textbf{44.3} & \textbf{39.0} & \textbf{51.1}~\gbf{+3.3} \\
    \bottomrule
\end{tabular}}
\end{center}
\vskip -5ex
\end{table*}

\begin{table*}[h]
\renewcommand\arraystretch{1.2}
\scriptsize
\setlength{\tabcolsep}{7pt}
\begin{center}
\caption{Per-class results on the \texttt{val} set of DeepAccident-360 dataset.}
\vskip -2ex
\label{tab:result:deepaccident}

\setlength{\tabcolsep}{2pt}
\resizebox{0.99\textwidth}{!}{
\begin{tabular}{l|c|ccccccccccccccccc|l}
    \toprule
    Method & \#Param. & \rotatebox{60}{Static} & \rotatebox{60}{Dynamic} & \rotatebox{60}{Building} & \rotatebox{60}{Fence} & \rotatebox{60}{Water} & \rotatebox{60}{Terrain} & \rotatebox{60}{Pedestrian} & \rotatebox{60}{Pole} & \rotatebox{60}{Road\_Line} & \rotatebox{60}{Road} & \rotatebox{60}{Side\_Walk} & \rotatebox{60}{Vegetation} & \rotatebox{60}{Vehicles} & \rotatebox{60}{Wall} & \rotatebox{60}{Ground} & \rotatebox{60}{Bridge} & \rotatebox{60}{Guard\_Rail} & \rotatebox{60}{Mean} \\
    \midrule
    HDMapNet & 12.8M & 12.7 & 15.4 & 32.1 & 9.2 & 26.7 & 65.8 & 0.0 & 2.3 & 14.1 & 73.4 & 64.9 & 46.1 & 15.2 & 16.9 & 58.0 & 1.4 & 36.3 & 28.8 \\
    BEVFusion & 53.7M & 10.0 & 11.1 & 24.5 & 9.1 & 17.9 & 57.4 & 0.0 & 0.8 & 12.0 & 66.4 & 55.7 & 40.7 & 13.2 & 16.9 & 43.5 & 0.0 & 25.8 & 23.8 \\
    BEVSegformer & 52.8M & 13.6 & 17.6 & 34.7 & 11.0 & 42.6 & 63.7 & 0.0 & 2.7 & 13.9 & 71.6 & 62.7 & 49.2 & 22.8 & 20.0 & 56.4 & 0.5 & \textbf{40.9} & 30.8 \\
    BEVFormer & 42.9M & 10.5 & 11.6 & 29.2 & 8.3 & 20.0 & 57.5 & 0.0 & 2.7 & 12.0 & 69.0 & 60.2 & 42.9 & 19.7 & 17.2 & 50.5 & 0.0 & 31.9 & 26.1 \\
    360BEV & 32.M & 15.6 & 16.3 & 42.1 & 10.1 & 50.4 & 68 & 0.0 & \textbf{3.6} & 16.9 & 76.7 & 70.4 &50.6 & 29.5 & \textbf{21} & 67.4 & \textbf{5.4} & 39.6 & 34.3 \\
    \textbf{OneBEV} (Ours) & \textbf{31M} & \textbf{17.5} & \textbf{19.6} & \textbf{46.7} & \textbf{11.5} & \textbf{59.5} & \textbf{72.5} & 0.0 & 2.8 & \textbf{19.2} & \textbf{80.1} & \textbf{73.5} & \textbf{51.3} & \textbf{30.4} & 19.7 & \textbf{73.9} & 4.7 & 31.7 & \textbf{36.1}~\gbf{+1.8}\\
    \bottomrule
\end{tabular}}
\vspace{-20pt}
\end{center}
\end{table*}

\noindent\textbf{Results on DeepAccident-360 Dataset.} 
The analysis of the results in Table~\ref{tab:result:deepaccident} indicates that OneBEV exhibits moderate performance improvements against 360BEV~\cite{teng2024360bev} on the DeepAccident-360 dataset, which is tailored to simulate real-world accident scenarios. Notably, it achieves a $+3.3\%$ improvement in \textit{Dynamic} IoU, $+4.6\%$ in \textit{Building} IoU, $+9.1\%$ in \textit{Water} IoU, $+4.5\%$ in \textit{Terrain} IoU, $+3.4\%$ in \textit{Road} IoU, $+3.1\%$ in \textit{Side Walk} IoU, $+6.5\%$ in \textit{Ground} IoU and $+1.8\%$ in all classes mIoU. The DeepAccident-360 dataset, being a synthetic dataset, presents unique challenges and opportunities, as it simulates complex and rare scenarios that are difficult to capture in real-world data. These results highlight OneBEV’s capability to handle diverse and complex environments, which is crucial for accurately simulating accident scenarios.

\subsection{Ablation Study}
\begin{table*}[]
\vspace{-4ex}
\begin{center}
\setlength{\tabcolsep}{2pt}
\small\centering
\caption{Results of ablation study for OneBEV on the nuScenes-360 \texttt{val} set. ‘VT’ denotes the method of view transformation. ‘\#Param.’ denotes the parameters of model, ‘\#Dims’ denotes embedding dimensions of model, ‘\#Layers’ denotes the repeat numbers of view transformation module. ‘\#Locs’ denotes number of sampling locations on the front view panoramic image. ‘\#Points’ denotes the number of all sampling points ‘\#Depth’ denotes the repeat number of vanilla VSS block.}
\label{tab:result:ablation}
\resizebox{0.99\textwidth}{!}{
\begin{tabular}{l|ccccccc|l}
    \toprule
     Backbone & VT & \#Param. & \#Dims & \#Layers & \#Locs & \#Points & \#Depth & mIoU \\
    \midrule
    MSCAN-B & MVT & 35.1M & 128 & 4 & 5 & 25 & 2 & 49.8\\
    VMamba-T & MVT & 31M & 128 & 4 & 5 & 25 & 2 & \textbf{51.1}~\gbf{+1.5}\\
    \midrule
    \midrule
    VMamba-T & CrossAtt & 28.8M & 128 & 2 & 6 & 24 & - & 35.8\\
    VMamba-T & MVT & 29.1M & 128 & 1 & 5 & 25 & 1 & \textbf{45.5}~\gbf{+9.7}\\
    \midrule
    \midrule
    VMamba-T & MVT & 29.1M & 128 & 1 & 5 & 25 & 1 & 45.5\\
    VMamba-T & MVT & 29.7M & 128 & 2 & 5 & 25 & 1 & 48.5\\
    VMamba-T & MVT & 30.7M & 128 & 4 & 5 & 25 & 1 & \textbf{50.5}~\gbf{+2.0}\\
    \midrule
    \midrule
    VMamba-T & MVT & 30.7M & 128 & 4 & 5 & 25 & 1 & 50.5\\
    VMamba-T & MVT & 31M & 128 & 4 & 5 & 25 & 2 & \textbf{51.1}~\gbf{+0.6}\\
    \midrule
    \midrule
    VMamba-T & MVT & 31M & 128 & 4 & 5 & 25 & 2 & \textbf{51.1}\\
    VMamba-T & MVT & 32.2M & 128 & 6 & 5 & 25 & 2 & 50.5~\obf{-0.6}\\
    \bottomrule
\end{tabular}}
\end{center}
\vspace{-4ex}
\end{table*}
\noindent\textbf{Effect of Backbone.}
Table~\ref{tab:result:ablation} (rows 1-2) compares the MSCAN-B~\cite{guo2022segnext} backbone in 360BEV~\cite{teng2024360bev} and the VMamba-T~\cite{liu2024vmamba} backbone within our model framework. Both use our MVT module for view transformation. The VMamba-T with the MVT module outperforms MSCAN-B in both model parameters and mIoU. This performance disparity is due to the architectural compatibility between VMamba-T and the MVT module, leading to more effective feature extraction and transformation. The VMamba-T backbone's design principles, such as better spatial resolution preservation and multi-scale feature aggregation, complement the MVT module's capabilities. Our findings suggest that the MVT module is more compatible with VMamba-T, highlighting the need for backbones designed to work effectively with advanced view transformation modules.

\noindent\textbf{Effect of View Transformation Module.}
Table~\ref{tab:result:ablation} (rows 3-4) shows an ablation experiment on the view transformation module. The model using multi-camera deformable cross attention~\cite{peng2023bevsegformer} performs poorly compared to the VMamba-T~\cite{liu2024vmamba} with the MVT module, despite the latter having a $0.3$M higher parameter count and achieving a $9.7\%$ higher mIoU. The cross-attention mechanism struggles to combine the complex features from the VMamba-T backbone, while the MVT module, designed for these features, ensures high-quality feature preservation and fusion. The increased parameter count allows for more complex computations, leading to better performance. These findings emphasize the importance of using view transformation mechanisms aligned with the backbone's feature characteristics for optimal performance.

\noindent\textbf{Effect of Hyperparameters.}
Table~\ref{tab:result:ablation} (rows 5-7) examines the impact of the number of layers in the MVT module. The model performs best with four layers, balancing feature transformation and computational complexity. Additionally, Table~\ref{tab:result:ablation} (rows 8-10) compares different depths of the vanilla VSS block~\cite{liu2024vmamba}, showing that greater depth leads to higher mIoU due to enhanced feature interaction. However, rows 11-12 indicate that increasing layers beyond six results in performance decline, suggesting overfitting. These findings highlight the need for careful hyperparameter selection to balance model complexity and performance, avoiding overfitting while ensuring robust feature extraction and transformation. Future work should explore adaptive mechanisms to optimize layer and depth configurations dynamically.

\section{Conclusion}
In this work, we propose a novel approach for BEV semantic mapping, termed OneBEV, which employs a single panoramic image as input to tackle challenges occurring in traditional multi-camera setups, including calibration inaccuracies, synchronization issues, and high computational complexity. Instead of leveraging traditional attention mechanisms, we introduce MVT that facilitates interaction between the front-view feature map and the BEV feature map, to alleviate spatial distortions inherent in panoramas. Extensive experiments showcase that OneBEV achieves superior accuracy with fewer parameters, demonstrating its efficiency and effectiveness across multiple tasks. We also contribute two datasets, nuScenes-360 and DeepAccident-360 to further facilitate panoramic-to-BEV semantic mapping research in the community. This paper aims to serve as a starting point, inspiring further discussions and developments on the panoramic-to-BEV semantic mapping in different domains, e.g., indoor navigation, robotic perception and augmented reality.
\section*{Acknowledgements}
\sloppy This work was supported in part by the Ministry of Science, Research and the Arts of Baden-Württemberg (MWK) through the Cooperative Graduate School Accessibility through AI-based Assistive Technology (KATE) under Grant BW6-03, in part by BMBF through a fellowship within the IFI programme of DAAD, in part by the InnovationCampus Future Mobility funded by the Baden-Württemberg Ministry of Science, Research and the Arts. We thank the Helmholtz Association Initiative and Networking Fund on the HAICORE@KIT and HOREKA@KIT partition.

\small
\bibliographystyle{splncs}
\bibliography{main}

\clearpage
\appendix
\section{Camera Model}
\begin{figure}[]
\vspace{-20pt}
\centering
\includegraphics[width=0.8\linewidth]{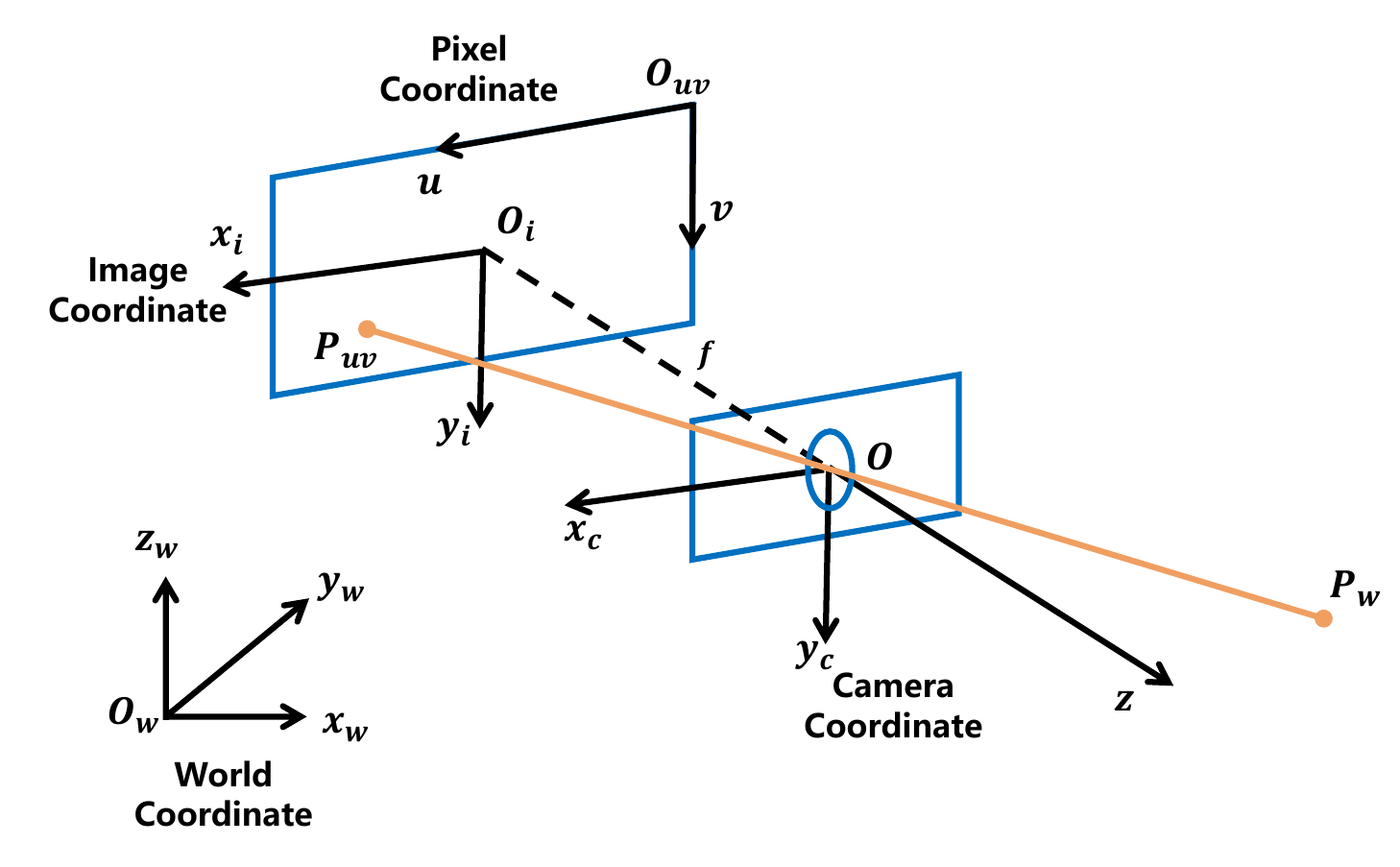}
\caption{\textbf{Camera model}. In the real world, a ray passes through the optical center $O$ of the camera from a point $P_w$ to a point $P_{uv}$ on the image plane.}
\label{fig:camera_model}
\vspace{-40pt}
\end{figure}

The camera model is a fundamental concept in computer vision and photogrammetry, describing the transformation from a three-dimensional (3D) world point to a two-dimensional (2D) image point. This process involves several key components: pinhole imaging, world coordinate system, camera coordinate system, image coordinate system, homogeneous coordinates and the mapping of 3D points to the 2D image plane, along with the intrinsic and extrinsic parameters of the camera. The specific structure of the camera model is shown in Fig.~\ref{fig:camera_model}.

\subsection{Pinhole Imaging} 
The pinhole camera model is a simplified version of a real camera, where a single small aperture (pinhole) projects light from the 3D world onto a 2D image plane. The pinhole camera can be described by a projection matrix that transforms 3D coordinates into 2D image coordinates.

\subsection{World Coordinate System} 
The world coordinate system is a global 3D Cartesian coordinate system that serves as a reference for all objects and cameras in the scene. It provides a universal frame of reference where each object's position and orientation are defined. The origin, orientation, and scale of the world coordinate system can be chosen arbitrarily based on the specific application or environment.

\subsection{Camera Coordinate System} 
The camera coordinate system is a 3D Cartesian coordinate system with its origin at the optical center of the camera. The $z$-axis points along the optical axis of the camera, the $x$-axis is horizontal, and the $y$-axis is vertical, forming a right-handed coordinate system.

\subsection{Image- and Pixel Coordinate System} 
\noindent\textbf{Image Coordinate System}:
The image coordinate system is a 2D Cartesian coordinate system defined on the image plane. The origin is typically located at the top-left corner of the image, with the $x$-axis pointing right and the $y$-axis pointing down.

\noindent\textbf{Pixel Coordinate System}:
The pixel coordinate system is a discrete 2D Cartesian coordinate system used to reference the position of pixels in a digital image. The origin of this coordinate system is typically located at the top-left corner of the image. The $x$-axis runs horizontally to the right, and the $y$-axis runs vertically downward. Each pixel's position is identified by its integer coordinates $(u, v)$, where $u$ denotes the horizontal position and $v$ denotes the vertical position. 

\subsection{Homogeneous Coordinates} 
In order to handle the transformations and projections in a more convenient mathematical framework, homogeneous coordinates are used. A point in 3D space is represented as $[X,Y,Z,1]^T$ in homogeneous coordinates, and a point in 2D image space is represented as $[u,v,1]^T$.

\subsection{Mapping 3D Points to 2D Image Plane} 
The transformation from 3D world coordinates $P_w:(X_w,Y_w,Z_w)$ to 2D image coordinates $P_{uv}:(u,v)$ is achieved through two steps involving both intrinsic and extrinsic parameters:
\begin{compactitem}
	\item World to Camera Coordinates: First, the 3D world coordinates are transformed into the camera coordinate system using the extrinsic parameters. This transformation is represented as:
    \begin{equation}\label{eq:world_to_cam}
    \begin{bmatrix}
      X_c \\
      Y_c\\
      Z_c\\
      1
    \end{bmatrix}
    = \begin{bmatrix}
      \mathbf{R} & \mathbf{t}\\
      \mathbf{0}^T & 1\\
    \end{bmatrix}
    \begin{bmatrix}
      X_w\\
      Y_w\\
      Z_w\\
      1
    \end{bmatrix}
    \end{equation}

    where $\begin{bmatrix}\mathbf{R} & \mathbf{t}\\0 & 1\\ \end{bmatrix}$ is the extrinsic matrix consisting of the rotation matrix $\mathbf{R}$ and the translation vector $t$.
	\item Projection to Image Plane: Next, the 3D coordinates in the camera system are projected onto the 2D image plane using the pinhole camera model. This projection is represented by the intrinsic matrix $\mathbf{K}$:
    \begin{equation}\label{eq:cam_to_image}
    s\begin{bmatrix}
      u \\
      v\\
      1
    \end{bmatrix}
    = \mathbf{K}
    \begin{bmatrix}
      X_c\\
      Y_c\\
      Z_c
    \end{bmatrix},
    \mathbf{K} = \begin{bmatrix}
    f_x & \gamma & c_x\\
    0 &  f_y & c_y \\
    0 &  0 & 1
    \end{bmatrix}
    \end{equation}
    
    Here, $s$ is a scaling factor, $f_x$ and $f_y$ are focal length in the $x$ and $y$ directions, $(c_x, c_y)$ is the principal point, and $\gamma$ is the skew coefficient.
\end{compactitem}

\section{Future Work}
In Table.~\ref{tab:flops}, we provide a detailed comparison of the computational cost between 360BEV and OneBEV, focusing on their components. The view transformation module in OneBEV is significantly more expensive (93.9 GFLOPs vs. 7.5 GFLOPs). Exploring an efficient view transformation remains for our future work.

\vspace{-4ex}
\begin{table*}[]
\begin{center}
\setlength{\tabcolsep}{2pt}
\small\centering
\caption{FLOPs Comparison.}
\label{tab:flops}
\resizebox{0.9\columnwidth}{!}{
\begin{tabular}{c|c|cccc}
\toprule
\textbf{Method} & \textbf{\makecell[c]{Total \\ FLOPs(G)}} & \textbf{\makecell[c]{Backbone \\ FLOPs(G)}} & \textbf{\makecell[c]{Neck \\ FLOPs(G)}} & \textbf{\makecell[c]{View Trans. \\ FLOPs(G)}} & \textbf{\makecell[c]{Head \\ FLOPs(G)}}\\ 
\midrule
360BEV & 107 & 61 & 1.5 & 7.5 & 37.1  \\
OneBEV & 177 & 44.7 & 1.7 & 93.9 & 37.1 \\ 
\bottomrule
\end{tabular}}
\end{center}
\end{table*}

\end{document}